\documentclass[letterpaper, 10 pt, conference]{ieeeconf}  %

\IEEEoverridecommandlockouts

\usepackage{times} %
\usepackage{amsmath} %
\usepackage{amssymb}  %
\usepackage{cite}
\usepackage{footnote}
\usepackage{comment}
\makesavenoteenv{tabular}
\usepackage[table,xcdraw]{xcolor}

\usepackage{multirow}
\usepackage{array}
\usepackage[T1]{fontenc}
\usepackage[utf8]{inputenc}
\usepackage{tabularx,ragged2e,booktabs}
\newcolumntype{C}[1]{>{\Centering}m{#1}}

\usepackage{makecell}
\usepackage[table,xcdraw]{xcolor}

\usepackage{nomencl}
\usepackage{relsize}
\usepackage{pdfpages}
\usepackage{caption}

\usepackage{color}
\usepackage{color, soul}
\usepackage{gensymb}
\usepackage{blindtext}
\usepackage{textcomp}
\usepackage{lipsum}

\usepackage{xcolor}

\makeatletter
\let\NAT@parse\undefined
\makeatother
\usepackage[hidelinks]{hyperref}
\usepackage{color}
\usepackage{wasysym}
\usepackage{subfig}
\usepackage{graphicx}
\usepackage{soul}
\usepackage{float}
\usepackage{amssymb}

\title{\LARGE \bf
AeroBuoy: A Drone Deployable, 3D Printed, Autonomous Robotic Buoy for Environmental Inspection in Remote and Hazardous River Systems}

\author{Reuben O'Brien*, Angus Lynch*, Minas Liarokapis %
\thanks{$^{*}$These authors have contributed equally to this work.}%
\thanks{Reuben O'Brien, Angus Lynch, and Minas Liarokapis are with the New Dexterity Research Group, Department of Mechanical and Mechatronics Engineering, The University of Auckland, New Zealand.}
\thanks{E-mails: robr763@aucklanduni.ac.nz, alyn649@aucklanduni.ac.nz
, minas.liarokapis@auckland.ac.nz,}%
}

\usepackage{eso-pic}
\begin{document}
\AddToShipoutPictureBG*{\AtPageLowerLeft{\makebox[\paperwidth][c]{\raisebox{0.22in}{\parbox[b]{\textwidth}{\scriptsize\centering \copyright~2025 IEEE. Personal use of this material is permitted. Permission from IEEE must be obtained for all other uses, in any current or future media, including reprinting/republishing this material for advertising or promotional purposes, creating new collective works, for resale or redistribution to servers or lists, or reuse of any copyrighted component of this work in other works. Accepted version. DOI: 10.1109/IROS60139.2025.11247654}}}}}

\maketitle

\begin{abstract}
Monitoring of waterways such as remote and hazardous rivers and streams is important so as to assess the impact of external factors including construction runoff or climate change. Versatile, autonomous robotic boats can offer excellent environmental inspection and monitoring solutions for remote, dangerous, or access protected water bodies but they have several shortcomings in terms of maneuverability. This paper proposes an environmental inspection system consisting of an autonomous data collection buoy which is designed to be deployed to inaccessible river systems using a drone. The system can perform a drop off and pickup of the buoy depending on the requirements of a particular location and monitoring task. Utilising the natural flow of the river the buoy autonomously steers down, using GPS and magnetometers so as to maintain the desired trajectory. The buoy is capable of measuring water temperature but it can also be equipped with a range of sensors such as water oxygen meter, sonar for river bed inspection, or turbidity for water clarity. This paper describes the system design, presents an analysis of the self-righting capabilities of the buoy, and shows a full system demonstration at the Ōrewa River in Auckland, New Zealand.
\end{abstract}
\section{Introduction}\label{sec:introduction}

Environmental monitoring in remote and dangerous rivers and other aquatic environments is a critical task for which robotic systems can provide a reliable, low-cost solution. Traditional methods, such as deploying boats or canoes for data collection or water sampling, present numerous challenges. These conventional platforms are often oversized relative to the narrow and winding nature of rivers, making them difficult to maneuver, particularly in confined spaces and turbulent waters. Additionally, their directional design can limit their ability to navigate sharp turns or reverse course, and they are highly susceptible to becoming entangled with floating debris and aquatic plants.

\begin{figure}[ht]
    \centering
    \includegraphics[width=\columnwidth]{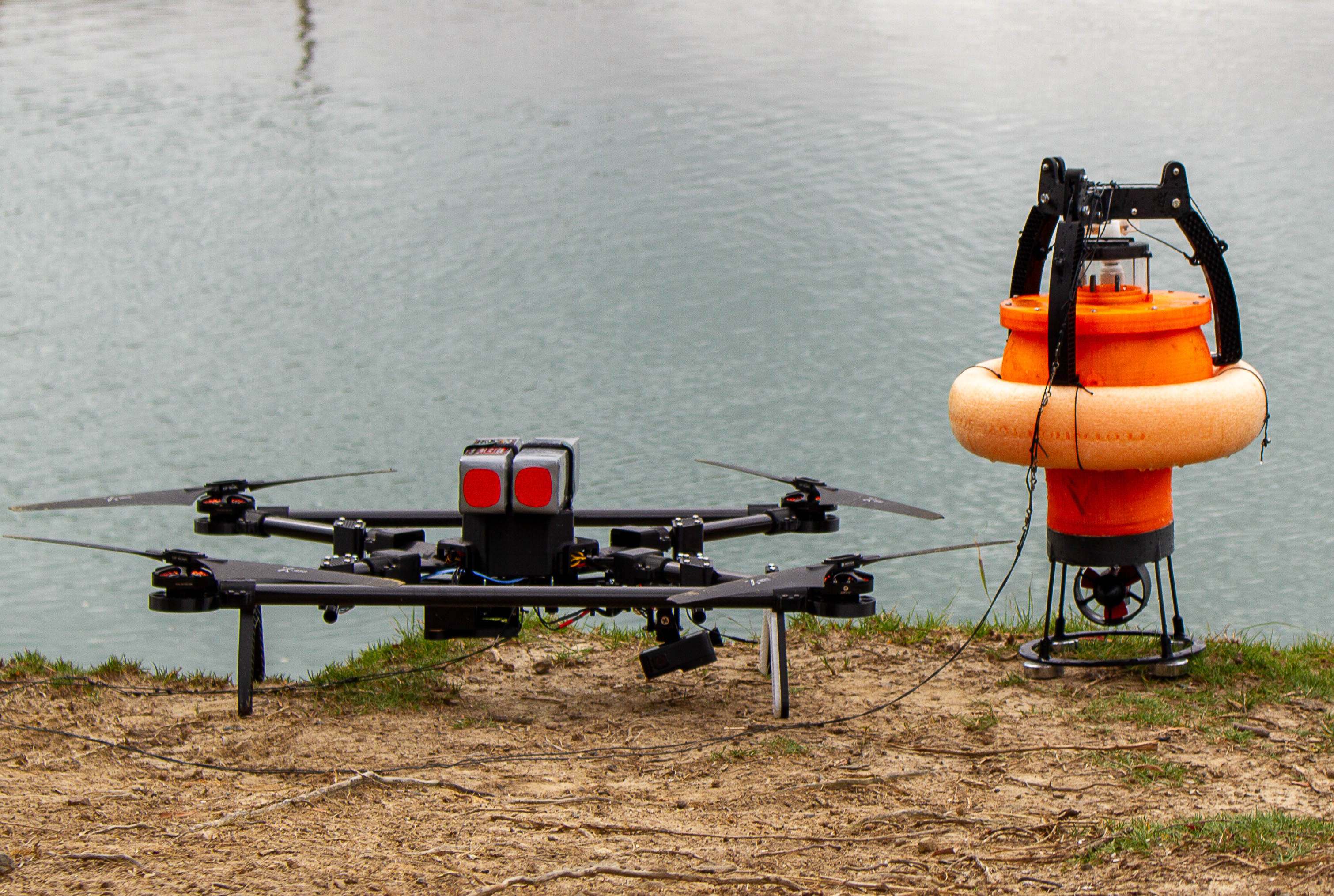}
    \caption{The proposed autonomous, 3D printed, robotic buoy `AeroBuoy' and the quadrotor drone that has been designed for the AeroBuoy deployment and collection in river systems.}
    \label{fig:cover}
\end{figure}

To mitigate some of these issues, jetboats have become a common choice for river navigation. Jetboats, with their impeller-based propulsion systems, are capable of operating in shallow waters and executing tighter turns. However, even these specialized vessels are not without their drawbacks. Their size and operational characteristics can still be suboptimal for the complex and dynamic environments often encountered in rivers, where factors such as shallow areas, debris, and rapid currents can pose significant risks.
The need for more efficient, adaptable, and resilient platforms for environmental inspections is evident, especially in locations where traditional methods are impractical or unsafe. The development of such systems can offer new possibilities for routine monitoring and rapid deployment in response to emerging environmental concerns. Furthermore, as rivers are often difficult to access, particularly in remote or hazardous regions, the ability to deploy and retrieve monitoring systems remotely would be highly advantageous. This capability would not only enhance the safety and efficiency of environmental monitoring efforts but also provide a consistent stream of valuable data over time.

This paper presents a novel 3D-printed, autonomous robotic buoy (Fig. \ref{fig:cover}) for environmental monitoring in inaccessible rivers. Its key novelty lies in the combined functionality of drone-based deployment/retrieval on both water and land along with a robust, self-righting design, enabling operation in debris-ridden waters. The buoy autonomously traverses downstream, driven by the river's flow, collecting temperature, imagery, and flow data. In this paper the design and experimental validation of the system is detailed.
\section{Related Work}\label{sec:related}
Environmental monitoring using autonomous systems has gained traction, particularly in remote and hazardous locations. Unmanned surface vehicles (USVs) and autonomous drones have emerged as viable platforms for gathering data in inaccessible environments. According to \cite{Jain2021}, applications for these technologies span seven categories: rescue, military, patrolling, disaster relief, environmental monitoring, inspection, and water sports rescue. Their primary benefits include reduced human involvement, faster response times, increased flexibility, and the ability to cover vast areas.
Several fixed autonomous systems have been developed to monitor environmental conditions. For example, \cite{Shukla2023} describes a stationary, moored buoy designed for continuous water quality monitoring. This system includes sensors to ensure safe drinking water standards are maintained in rivers. Despite its effectiveness, stationary solutions like this are constrained by their inability to move, making them less suitable for large or dynamic environments like rivers. Larger mobile systems have also been explored for monitoring aquatic environments, in \cite{Si2023}, a USV raft capable of traversing oceanic environments while measuring various parameters at different speeds was presented. These larger systems, although effective in open waters, are less suited for narrower or obstacle-laden rivers due to their size and limited manoeuvrability. The propulsion systems, like wave gliders \cite{hine2009}, which rely on ocean waves for long-distance travel, also fall short in river environments where wave energy is minimal. Smaller, motorized buoy systems have shown promise in autonomous monitoring. For example, in \cite{Guan2022} a motorized buoy with dual thrusters was developed that optimizes path planning and manoeuvrability. However, even with improved efficiency, systems designed for calm lakes may struggle in dynamic and debris-filled river environments. This is where leveraging natural river currents could significantly reduce energy consumption. MIT designed and built the `BoB' system offering a significant innovation in the buoy movement. Designed for omnidirectional travel, this 7.4 kg system uses three points of propulsion to achieve high stability in calm waters \cite{Zoss2018}. It was later upgraded to be a 3.3 kg system \cite{Kwa2020}. Although it shows promise in lake environments, its application to river systems remains limited by its inability to navigate turbulent waters.

Research on autonomous surface vehicles (ASVs) specifically designed for rivers is ongoing. Systems such as \cite{madeo2020low} provide a low cost, monitoring system, but it is limited by its size and weight. In \cite{Schneider2016}, the authors proposed a method for using ultrasonic sensors to help ASVs monitor oil spills while avoiding obstacles like rocks and branches. This sensor-based navigation is critical for buoys and drones in river environments, where river currents \cite{sanjou2021portable} and debris create unpredictable challenges, within the river boundaries \cite{zhang2023river}. Drone integration with buoys has gained attention as an effective deployment strategy in remote or hazardous areas. In \cite{Franek2023} a stationary monitoring buoy that is deployable and retrievable by a drone was demonstrated. It featured floats for safe water landings, showcasing the potential for drone-based buoy deployment in dynamic environments. However, other drone systems, like \cite{Ajgaonkar2020}, which deployed lifebuoys using a servo mechanism, lacked retrieval capabilities and required manual loading. Simulated dynamic models for a tethered buoy and drone are presented in, \cite{Kourani2021}, \cite{kourani2021bidirectional}, showing potential in both calm and wave conditions. Real-world testing of a similar system has been done in `The Bathy Drone' \cite{Diaz2022}, which demonstrates a tethered drone system capable of bathymetry. The sonar is attached to a multi-hull boat that is towed over open water.

Unmanned Aerial Vehicles (UAV's) or `drones' have been developed to have many different ways of implementing aerial grasping and manipulation. In \cite{aerial-grasp-review}, the authors review many of these methods, involving all kinds of mechanisms and actuation methods. In \cite{lynch2024}, a grasping mechanism on a drone is implemented using a dynamixel motor, where the drone uses it's grasper to perch onto a power line. In \cite{6385963} a system where multiple drones support a net using tethers mounted to the underside of the drone is shown. \cite{10018625} presents a drone which is able to lower a net based mechanism, suspended from a cable below the drone to pickup packages for delivery. The mechanism is actuated by motors mounted onto the net itself, so would be unsuitable to pickup or drop off into water.\cite{9650012} presents the design of a three-fingered claw mechanism installed on a drone for picking up objects, particularly debris from power lines. A similar approach could be adapted for aquatic buoy retrieval by integrating a larger tether, ensuring the drone remains at a safe distance from the water.

This paper proposes a novel approach to addressing the challenges of river-based environmental monitoring. Section III delves into the design of the proposed system, focusing on its novel aspects. Section IV discusses the experimental validation of the platform in the Ōrewa River and presents the results. Finally, Section V concludes the paper, discussing the potential impact of this solution on future environmental monitoring efforts.

\begin{figure*}[ht]
    \centering
    \includegraphics[width=0.9\textwidth]{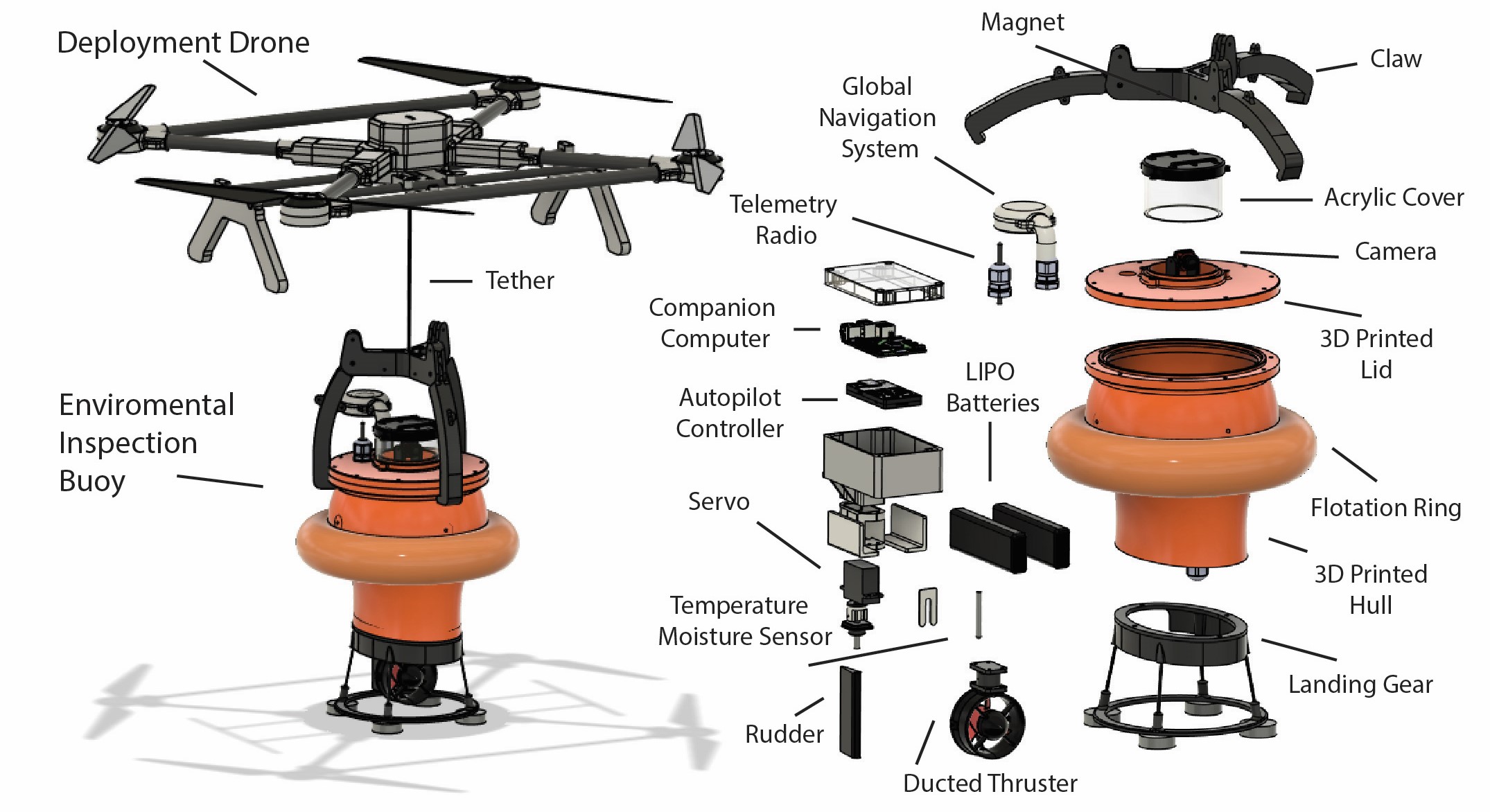}
    \caption{Exploded view of AeroBuoy, the proposed drone deployable, autonomous robotic buoy, showing all components used to implement it, as well as the overall design with the drone connected to the buoy via it’s tether and claw.}
    \label{fig:design-explode}
\end{figure*}

\begin{figure}[ht!]
    \centering
    \includegraphics[width=0.9\columnwidth]{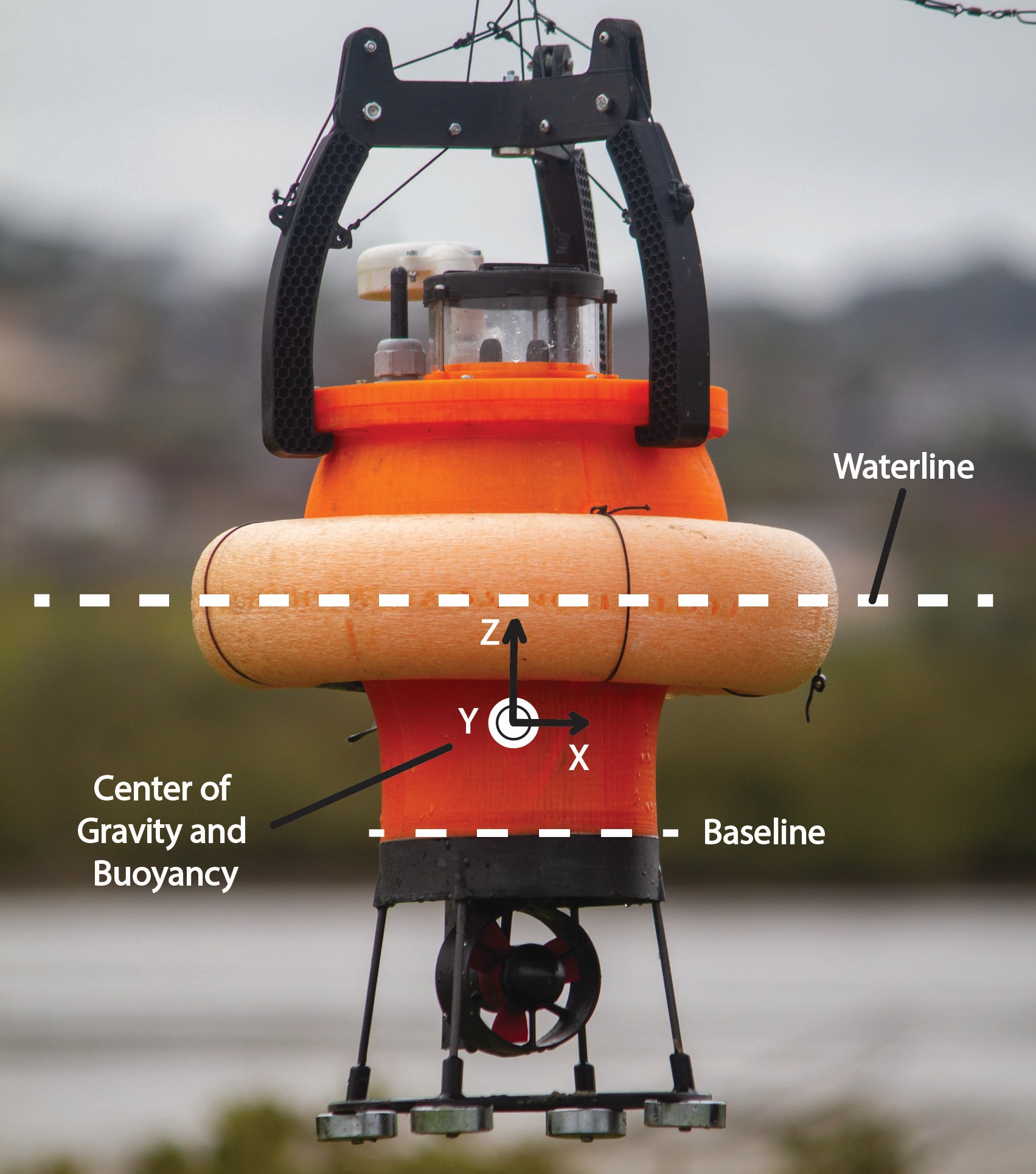}
    \caption{The AeroBuoy is made up of a 200 mm tall and wide hull (400 mm tall including peripherals). From the baseline the waterline is located 105 mm in the Z direction. Center of gravity and buoyancy have been designed to be located at the same position in all axes at 64 mm above the baseline. This provides the highest level of stability and control.}
    \label{fig:waterline-buoy}
\end{figure}

\section{Design \& Modelling}\label{sec:design}
\subsection{Buoy Design}

The environmental inspection AeroBuoy is comprised of one main hull shown in Fig. \ref{fig:design-explode}, that is 3D printed from polyethylene terephthalate glycol (PETG) as discussed in \cite{obrien2024}. A flotation ring is mounted around it and attached via four external mounting points to provide greater stability while ensuring that the buoy will always float regardless of any damage to the main hull. Most importantly it acts as a barrier to rocks, branches, or other obstacles that system might unexpectedly collide with, damping the impact to the electronics and protecting the hull at the waterline. The lid is sealed using a custom sized foam o-ring that seals from the compression of the 10 flange bolts around the rim. To allow for grasping the lid and flange are circular and reinforced with a 10 mm overhang for the claws to hook onto. On top of the lid is an enclosed camera which can either be a piloted first person view (FPV) or a logger connected to the companion computer. A global positioning system (GPS) is mounted alongside the telemetry radio with no physical obstructions. This allows clear communication to an off-board computer and strong satellite communication. Above the camera's lid two mild steel strips that have been coated with paint and allow for the claw's magnet to locate itself centrally.
To provide a stable landing and take off, a 3D printed base with connecting carbon rods provide the standoff. Additionally, the landing gear acts as protection to the propulsion system which is a ducted propeller and a rudder that is controlled via a double o-ring sealed shaft. Speeds of up to 1.5 m/s can be achieved in steady water with operation possible against river flows of up to 1 m/s. Water quality sensors are also protected with the temperature and moisture probe extending through the base of the hull. Internally, the system is powered by two 2.6 amp hour, lithium-polymer batteries that power the propulsion, navigation, and sensing electronics. They are able to provide up to one hour of run time at steady speeds of 0.5 m/s. Utilising the flow of the river only corrections towards the center of the river are needed, therefore just under two hours of operation is possible with the constant base draw of 3A. The onboard computing is stored within an IP68 box to act as the black box in case of catastrophic failure.

\subsection{Control and Sensing Hardware}

Since the AeroBuoy is deployed from land it should be able to detect when it is in or out of the water to begin autonomous navigation. This is achieved through the use of a moisture sensor that uses capacitance to determine dielectric permittivity of the surrounding surfaces, that being air or water. Once in water it allows the autopilot, shown in Fig. \ref{fig:design-explode}, to switch from manual to auto. Using the onboard software Ardupilot \cite{ardupilot}, which has configured to be a 'Rover' type system, it will navigate to the preset way-points. It is able to do this with the use of the the pixhawk GPS. The onboard companion computer is able to log on board humidity and temperature to ensure there are no leaks nor overheating of the propulsion or computing components. The system also provides a platform for more environmental sensors such as external water temperature, turbidity, PH level, and single point sonar echo-sounders can be linked to provide the measurements of water conditions as needed by the environmental study.

\subsection{Buoy Self Righting Analysis}

\begin{figure}[ht!]
    \centering
    \includegraphics[width=\columnwidth]{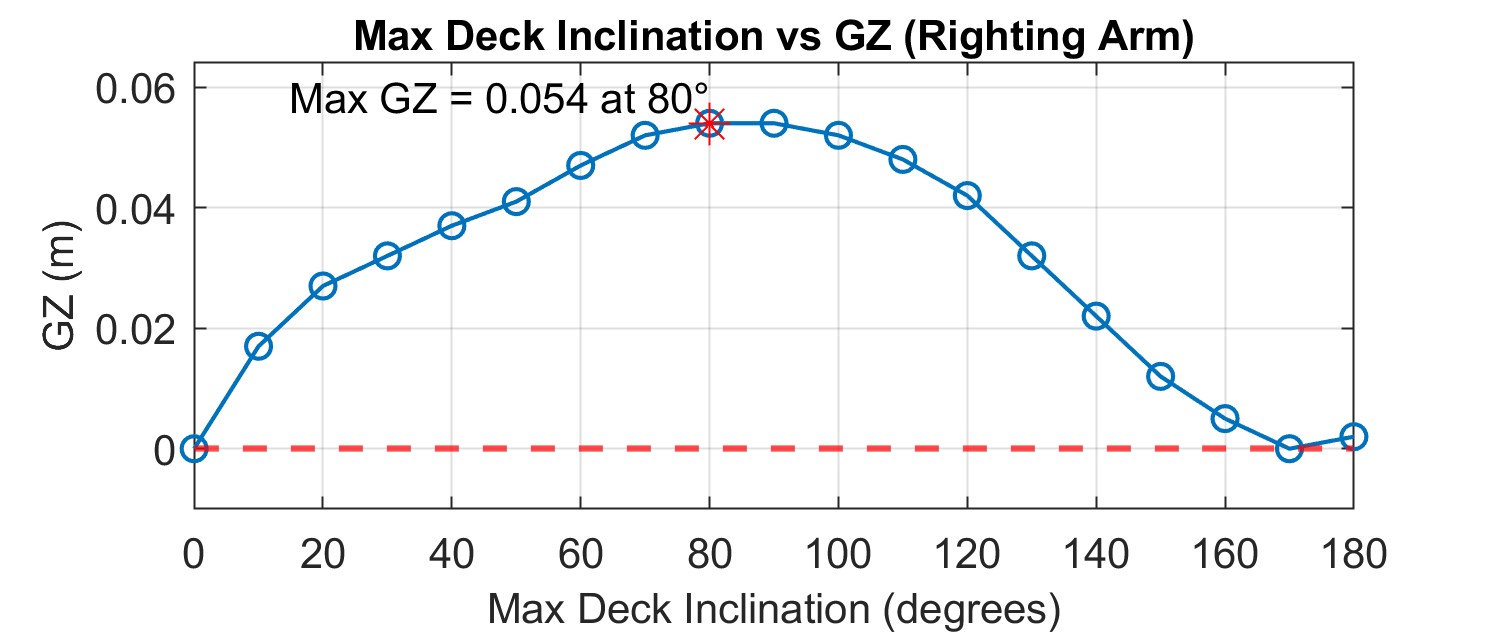}
    \caption{Righting arm of the buoy hull ranging from 0 degrees (upright) to 180 degrees (fully inverted). This highlights that regardless of the angle of the buoy in the water, there will always be a moment applied to rotate it back to level.}
    \label{fig:self-righting}
\end{figure}

To ensure optimal performance of the AeroBuoy, the hull was optimised and validated through simulation software MAXSURF. It provided a hydro-static and stability analysis of the design that allowed for optimal hull shapes and placement of components. For the system to work it needed to be self righting and stable through normal travel. In Fig. \ref{fig:self-righting}, there is no point of vanishing stability, the curve is always above zero. Therefore at no point will the system end up in a non-recoverable position. To achieve this four 100 g masses are mounted at the base of the landing gear to act as a ballast. Along with the batteries being mounted at the base of the hull. This brought the center of mass inline with the buoyancy at 64 mm above the base line, as seen in Fig. \ref{fig:waterline-buoy}. Overall weight of the hull and all the navigation and sensing electronics is 2.56 kg with the system becoming overloaded and therefore unstable at 5.5 kg. The waterline sits just under the mid-line of the flotation ring at 105 mm above the base-line. This adds a lever arm to counter the moment created by the propulsion during travel.

\subsection{Drone \& Claw Design}

\begin{figure}[h]
    \centering
    \includegraphics[width=0.8\columnwidth]{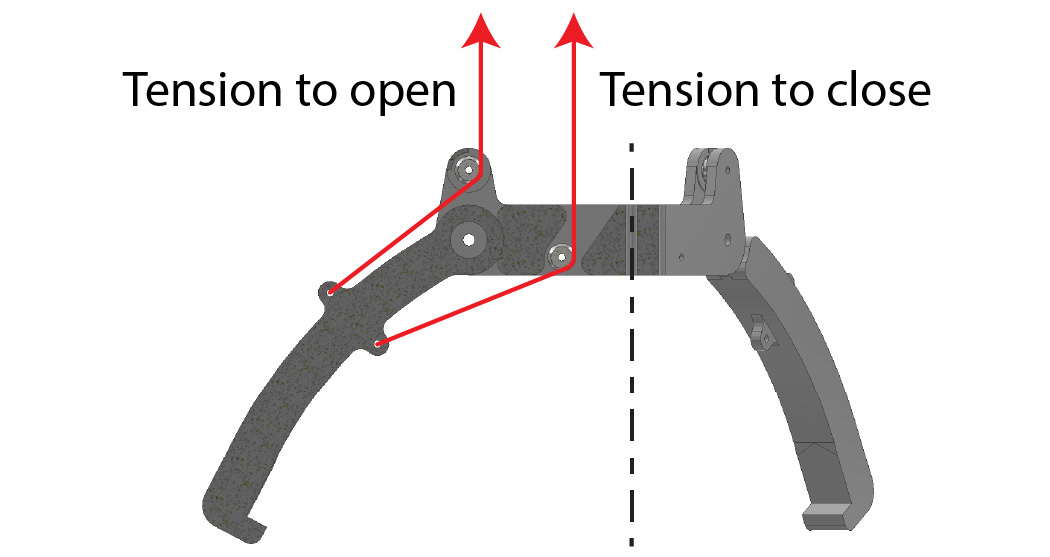}
    \caption{Actuation strategy of the cable driven claw mechanism used to pickup and drop off the AeroBuoy. }
    \label{fig:claw}
\end{figure}

The drone presented in this paper to be apart of the overall system is designed to lift large payloads, and is equipped with the actuation hardware needed to operate the claw mechanism. The drone is built using a PX4 based flight controller \cite{10.1109/icra.2015.7140074}, T-Motor MN5006 450KV motors and 18" propellers. This results in a drone with a takeoff mass of 3.6 kg, including the 1.5 kg of batteries, capable of lifting an extra 6.4 kg of payload. The drone is powered by a 9 amp hour, 6 series cell lithium-polymer battery, which allows for an estimated flight time of 28.2 minutes at the take off mass, and 12.6 minutes with the proposed 2.56 kg AeroBuoy system carried in the claw. The drone is fitted with a GNSS receiver to estimate  position, as well as an ArkFlow optical flow sensor which fuses with the global position estimate. This allows the drone to be more responsive to high frequency changes in position due to disturbances, such as the slung load of the buoy hanging beneath. The combination of these sensors allows for robust position control of the drone by the pilot. The drone needs to be able to actuate the claw when commanded to by the pilot to release or pickup the buoy. To achieve this, a Dynamixel XM430-350R servo motor is used, with a driver board that converts the PWM servo signal from the flight controller to the RS485 dynamixel commands. In Fig. \ref{fig:claw}, the actuation strategy of the cable driven claw mechanism is depicted. The idea is to have all of the actuation onboard the drone, allowing the claw to be totally submersible without consideration for waterproofing actuators at the end effector. Two main lines are between the claw and the drone, and at the claw end the main lines are spliced symmetrically onto each finger in the manner shown in Fig. \ref{fig:claw}. The 'close' cable is mounted to the center of the drone, passively closing the claw under it's on weight. To open the claw, tension is applied to the 'open' cable via a servo motor on the drone, taking up the weight of the claw and subsequently opening the grasp. A benefit of this strategy is that the claw's grasp is only improved by more mass, as the tension on the line is applied to closing the fingers. Another benefit is that the actuator on the drone only needs to be powerful enough to overcome friction between the buoy and the fingers, and no actuator action is needed to hold onto the buoy, increasing reliability. Finally, there is a magnet mounted to the center of the claw, which allows the claw to make an initial contact with the buoy and `lock on' in order to make the pickup maneuver more robust. After the drone starts to lift up, the magnetic connection separates as the fingers take up the weight of the buoy.

\begin{figure*}[ht]
    \centering
    \includegraphics[width=0.9\textwidth]{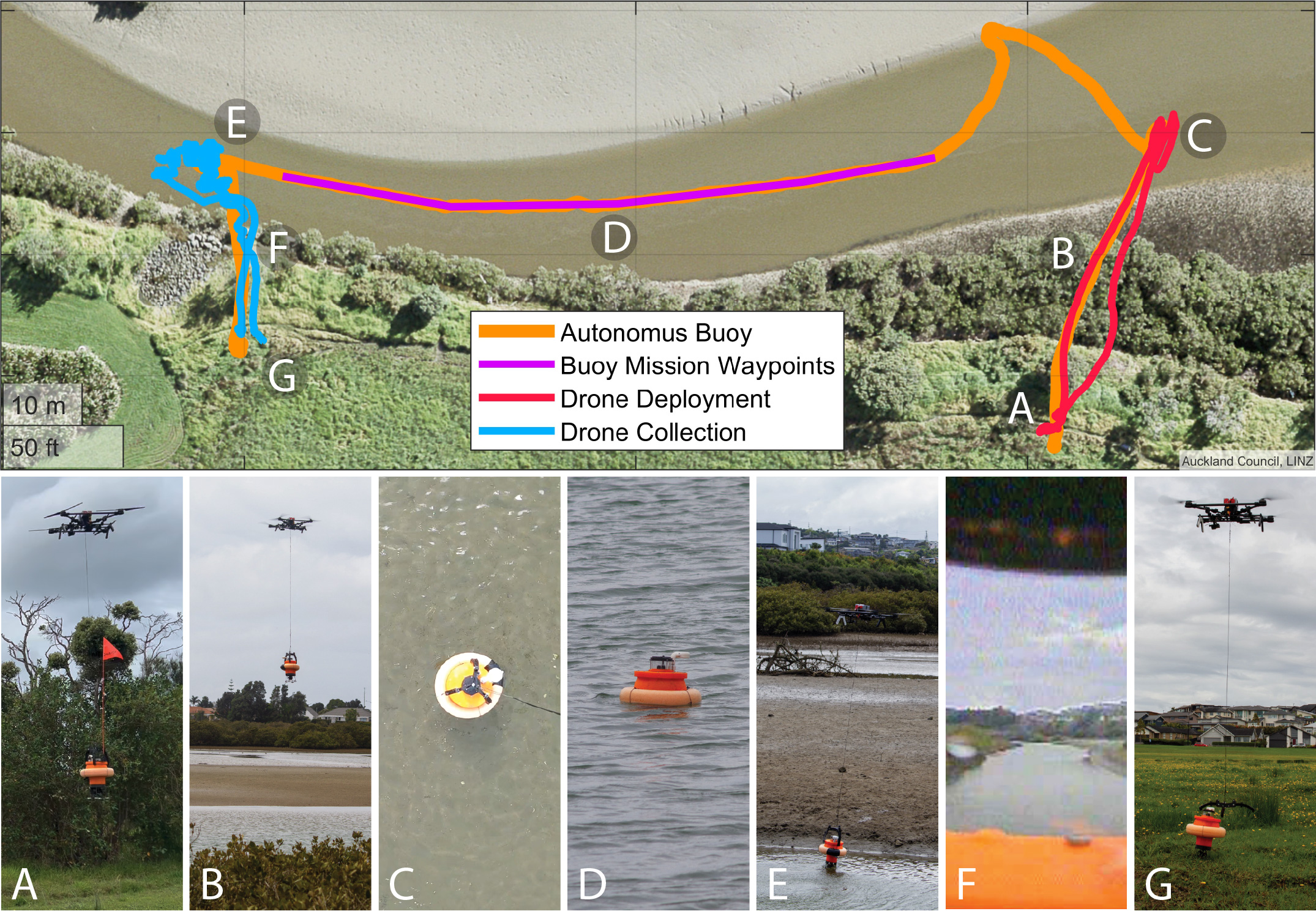}
    \caption{GPS tracking \& timeline images of the autonomous AeroBuoy deployment and retrieval throughout an experiment at the Ōrewa River in Auckland, New Zealand (36°35'45.7"S, 174°40'40.8"E). The buoy is picked up from the land by the drone (in A), traverses over inaccessible terrain (in B), and is deployed into the water (in C). The buoy conducts its autonomous data collection mission (in D), until it reaches the end where it autonomously loiters, and is collected again by the drone (in E). The drone carries it out of the water (in F) and returns the buoy to land for the data to be retrieved (in G).
    }
    \label{fig:experiment}
\end{figure*}

\section{Experiments}\label{sec:results}

\subsection{Deployment and Collection Experiment}
In order to validate the proposed system, an experiment was conducted that demonstrated all of the main aspects of the system. The site of the experiment was Ōrewa River (36°35'45.7"S, 174°40'40.8"E), a tidally affected river estuary, located in Auckland, New Zealand. At the time of the experiment presented in this work, the tide was incoming, and therefore the water flow through the river was from east to west. The river is difficult to access without traversing through bushes and mangroves, so taking environmental samples of the river would be difficult and time consuming, and if a study wished to sample many points along the river, this would become nearly impossible. In preparation for the experiment, a series of waypoints were planned using satellite imagery to create a path along the river, keeping the path approximately centered in the river. The waypoints had a 2 metre radius around them that was considered to have arrived at the waypoint and proceed to the next. The path between these waypoints is represented by the pink line in Fig. \ref{fig:experiment}. This mission plan is then uploaded wirelessly to the buoy via it's telemetry radio. Fig. \ref{fig:experiment} presents the results of this whole system experiment, where the GPS data from the drone and buoy are combined onto a single satellite map view, and the river and bush terrain are both visible. There are also 7 key points of the experiment, represented by A to G, with corresponding images from various view points available, such as on board the buoy, on board the drone, or afar as the observer. First, in A of Fig. \ref{fig:experiment}, the buoy is loaded into the claw of the drone and lifted into the air with the drone. The control of the drone is from the pilot who uses a combination of the onboard, downward facing camera, and line of sight of the drone. The drone then carries the buoy, shown in B, over the inaccessible terrain towards the starting point of the planned path. The pilot then lowers the altitude of the drone, shown in C, and once contact with the water is made, the pilot releases the claw. On contact with the water, the buoy enters into 'auto' mode where it begins its autonomous mission, shown in D. Once the buoy has completed the mission, it enters into 'loiter' mode. Loiter mode keeps the buoy stationary in the river, regardless of the flow of water. This allows the drone to collect the buoy from the river, shown in E. The pilot hovers over the buoy and aligns the claw with the top of the buoy. The magnet assists in this process as it can lock onto the top of the buoy. Once this has happened, the pilot can close the claw and lift it up out of the water. The buoy is then disarmed, and is carried out of the river by the drone, shown in F, and returned safely to the land, shown in G, where the data collected from the buoy can now be gathered, and the system is ready to do another deployment.

\begin{figure}[h]
   \centering
   \includegraphics[width=1\columnwidth]{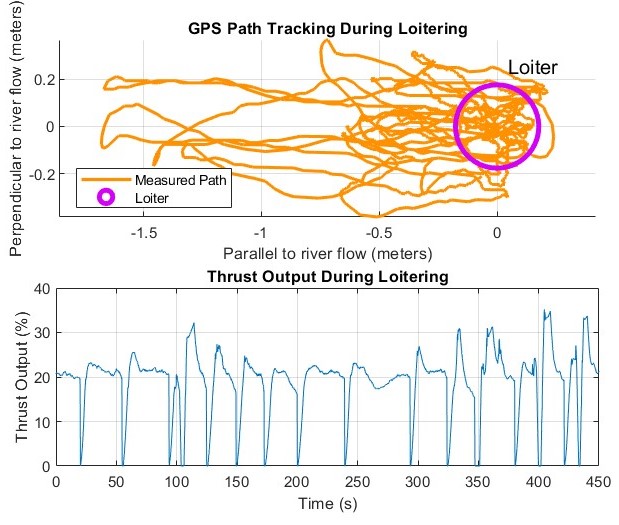}
   \caption{Loitering at a set pickup location for the drone under a \textasciitilde 0.5 m/s river flow. It was able to stay within 0.5 m perpendicular to the flow and 2 m parallel, with external disturbances from the drone. This was conducted over a 7.5 min sequence.}
   \label{fig:loiter}
\end{figure}

\subsection{Loitering During Pickup}
Once the AeroBuoy has traversed the set way-points it needs to hold position so that the drone can extract it. To do this, the buoy transitions to loiter mode where it will hold its GPS position. In the situation seen in Fig. \ref{fig:loiter} the buoy has to hold its position in a river that is flowing at 0.5 m/s. Its initial travel was with the flow of the river, but to hold position it re-orientates to face it and applies and average of 20\% throttle to maintain location as seen in Fig. \ref{fig:loiter}. During pickup the claw on the drone will act as an external disturbance that can move the buoy out of its desired loiter position. This can be either dragging it or rotating it. There were three major rotations where the buoy got pointed downstream and had to recover which lead to a shift of up to 1.75 m. During the rest of the 7.5 min duration it was able to maintain within a 0.5 m radius circle.
Alternative use cases could be tether-less monitoring in different locations in the river or even mooring-less monitoring out in the ocean like \cite{joe2014development}. For open water uses and in areas where other systems may be present a flag can be mounted to the lid that provides visibility to heights of 1.5 m above the water.

\subsection{Other Experiment Locations}
\begin{figure}[h]
    \centering
    \includegraphics[width=\columnwidth]{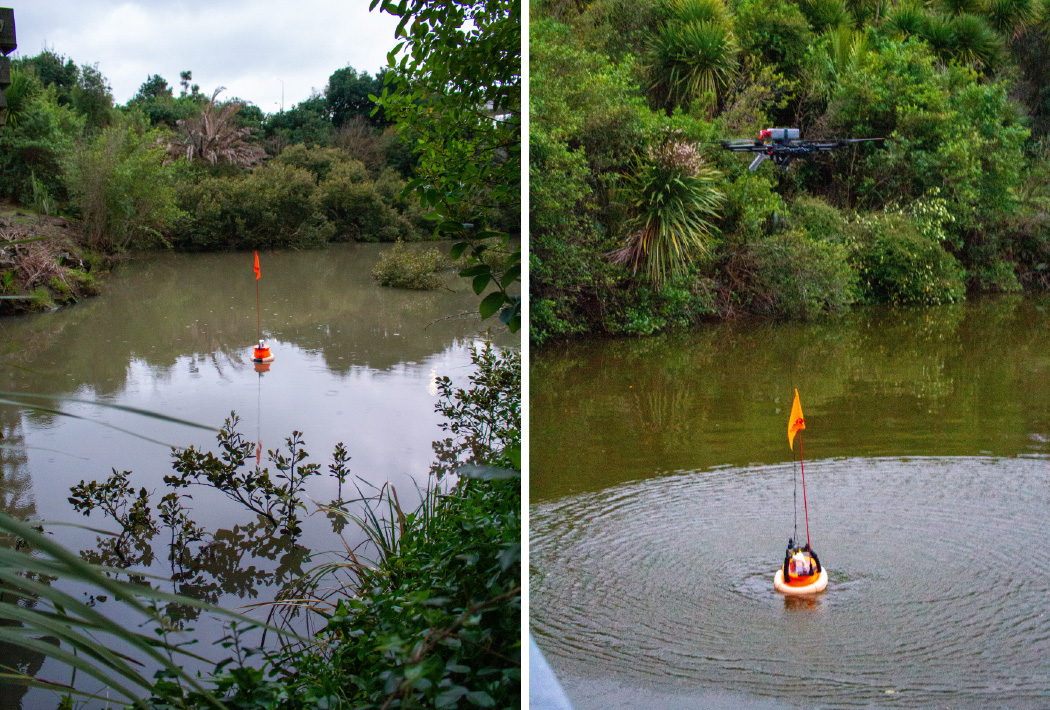}
    \caption{A demonstration of the capabilities of the proposed drone deployable, 3D printed, autonomous robotic buoy `AeroBuoy' at Oakley Creek (left) \& Newmarket Park Pond (right) in Auckland, New Zealand.}
    \label{fig:oakley}
\end{figure}

To ensure repeatability, other locations were trialed such as in Fig. \ref{fig:oakley}. In Oakley Creek the buoy was required to navigate around mangroves, fallen trees, and under low bridges. It performed all of this while it rained both with and against the natural river flow of 0.3 m/s. For the Newmarket Pond it was testing the ability to navigate through excessive plant growth where weeds, algae, and organic matter would clog the propulsion system. It was successful due to the ducted-thruster and the landing legs acting as a barrier for larger objects such as branches. During pickups, there are external disturbances of both the claw and the down thrust of the propellers. This can be seen on the right in Fig. \ref{fig:oakley} where the water is rippling around the buoy and causing it to drift.

\subsection{Experimental Discussion}
The proposed system shows promise in being able to deploy the AeroBuoy into remote, inaccessible locations through the use of the drone, utilising onboard video to the pilot. Traversing down the river with the natural flow of the water also shows significant promise. Further work is required with the pickup of the buoy due to the loitering not being pinpoint accurate, the claw moving the buoy around, and the down wash of the drone disturbing the buoy. Utilising a vision-based autonomous control system during pickup would simplify drone retrieval, while integrating Real-Time Kinematic Global Navigation Satellite System (RTK-GNSS) would enhance positional accuracy to below 3 cm \cite{stempfhuber2012precise}. For longer missions, equipping the buoy with an energy generation system such as solar panels would enhance its capability to operate and traverse extended sections of river. However, in its current design, the buoy remains efficient because it relies primarily on the natural flow of the water for movement.

\subsection{Video and Further Information}
A video presenting the experimental validation of the efficiency of the AeroBuoy as well as further information can be found at the following URL:
\begin{center}
\href{http://www.newdexterity.org/aerobuoy}{www.newdexterity.org/aerobuoy}
\end{center}

\section{Conclusions}\label{sec:conclusions}

Autonomous monitoring of waterways, especially remote and hazardous river systems is a challenge for robotic systems but also an opportunity as they can help assess the effect of external factors in these environments. In this work, we introduced an environmental inspection system featuring an autonomous data collection robotic buoy designed for deployment in inaccessible river systems via a drone. The system allows for both drop-off and retrieval of the buoy based on the specified location and monitoring requirements. Once deployed, the buoy autonomously navigates the river using GPS and magnetometers to follow a set course. While it is currently equipped with a water temperature sensor, it can also support other sensing modalities. We have presented the system’s designs, the self-righting capabilities of the buoy, and a full demonstration of the monitoring capabilities of the system conducted at the Orewa River in Auckland, New Zealand.

\bibliographystyle{IEEEtran}
\bibliography{REF}
\end{document}